\documentclass{article} 
\usepackage{iclr2026_conference,times}

\usepackage{amsmath,amsfonts,bm}

\def\eqref#1{equation~\ref{#1}}

\def\1{\bm{1}}

\DeclareMathAlphabet{\mathsfit}{\encodingdefault}{\sfdefault}{m}{sl}
\SetMathAlphabet{\mathsfit}{bold}{\encodingdefault}{\sfdefault}{bx}{n}

\usepackage{hyperref}
\usepackage{url}
\usepackage{microtype}
\usepackage{graphicx}
\usepackage{subcaption}
\usepackage{booktabs} 

\usepackage{hyperref}
\usepackage[utf8]{inputenc} 
\usepackage[T1]{fontenc}    
\usepackage{hyperref}       
\usepackage{url}            
\usepackage{booktabs}       
\usepackage{amsfonts}       
\usepackage{nicefrac}       
\usepackage{microtype} 
\usepackage{microtype}      
\usepackage{tikz}
\usetikzlibrary{positioning}
\usepackage{tikz}
\usepackage{float}
\usepackage{graphicx}
\usepackage{subcaption}
\usepackage{booktabs} 
\usepackage{multirow} 
\usepackage{xcolor}   
\usepackage{graphicx} 
\usepackage[table]{xcolor}
\usetikzlibrary{
    arrows.meta,       
    positioning,       
    shapes.multipart,  
    shapes.misc,       
    shapes.symbols,
    shapes.geometric,  
    calc,              
    fit,               
    backgrounds,       
    decorations.pathreplacing 
}
\usepackage{lipsum}
\usepackage{tabularx}
\usepackage{fancyhdr}       
\usepackage{graphicx}       
\usepackage{natbib}
\usepackage[most]{tcolorbox}
\usepackage{multirow}
\usepackage{threeparttable}
\graphicspath{{media/}}     
\usepackage{algorithm}
\usepackage{algorithmic}
\usepackage{amsmath}
\usepackage[capitalize,noabbrev]{cleveref}
\usepackage{wrapfig}
\usepackage{booktabs}
\usepackage{pifont}
\usepackage[most]{tcolorbox}
\tcbset{colback=gray!5, colframe=gray!40, rounded corners, boxrule=0.4pt}

\usepackage{amssymb}
\usepackage{mathtools}
\usepackage{amsthm}

\usepackage[capitalize,noabbrev]{cleveref}

\title{Learning Agent Routing From Early Experience}

\author{
\textbf{Yimin Wang}$^{*2,4}$ \hspace{1.2em}
\textbf{Jiahao Qiu}$^{*1}$ \hspace{1.2em}
\textbf{Xuan Qi}$^{3}$ \hspace{1.2em}
\textbf{Xinzhe Juan}$^{2,4}$ \hspace{1.2em}
\textbf{Jingzhe Shi}$^{3}$ \\[0.25em]
\textbf{Zelin Zhao}$^{6}$ \hspace{1.2em}
\textbf{Hongru Wang}$^{5}$ \hspace{1.2em}
\textbf{Shilong Liu}$^{\dagger 1}$ \hspace{1.2em}
\textbf{Mengdi Wang}$^{\dagger 1}$ \\[0.6em]
$^{1}$AI Lab, Princeton University \hspace{1.2em}
$^{2}$University of Michigan \\
$^{3}$Institute for Interdisciplinary Information Sciences (IIIS), Tsinghua University \\
$^{4}$Shanghai Jiao Tong University \hspace{1.2em}
$^{5}$University of Edinburgh \hspace{1.2em}
$^{6}$King’s College London
}

\begingroup

\footnotetext{$^*$ Equal contribution.}
\footnotetext{$^\dagger$ Corresponding authors.}
\endgroup
\iclrfinalcopy 
\begin{document}

\maketitle

\begin{abstract}
LLM agents achieve strong performance on complex reasoning tasks but incur high latency and compute cost. In practice, many queries fall within the capability boundary of cutting-edge LLMs and do not require full agent execution, making effective routing between LLMs and agents a key challenge. 
We study the problem of routing queries between lightweight LLM inference and full agent execution under realistic cold-start settings. 
To address this, we propose BoundaryRouter, a training-free routing framework that uses early behavioral experience and rubric-guided reasoning to decide whether to answer a query with direct LLM inference or escalate to an agent. BoundaryRouter builds a compact experience memory by executing both systems on a shared seed set and retrieves similar cases at inference time to guide routing decisions.
To evaluate this method, we introduce RouteBench, a benchmark covering in-domain, paraphrased, and out-of-domain route settings. Experiments show that BoundaryRouter reduces inference time by 60.6\% compared to the agent while improving performance by 28.6\% over direct LLM inference, outperforming prompt-based and retrieval-only routing by an average of 37.9\% and 8.2\%, respectively.
\end{abstract}

\begin{figure}[htbp]
\centering
\includegraphics[width=0.95\linewidth]{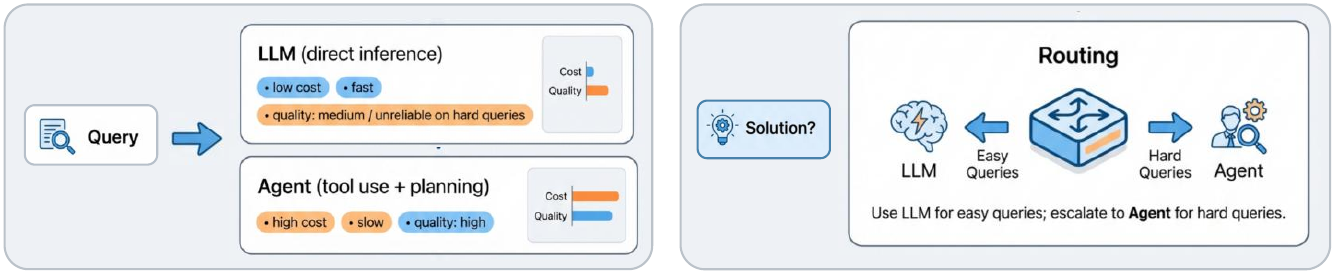}
    \caption{Motivation and overview of routing. Direct LLM inference is fast and low-cost but can be unreliable on harder queries, while full agent execution is slower and more expensive. A router dispatches each query to the appropriate system, using the LLM for easy cases and escalating to the agent when needed to achieve a better accuracy–latency trade-off.
    }
    \label{fig:motivation}
\vspace{-0.2in}
\end{figure}

\section{Introduction}
\label{Introduction}

Large language model agents have recently emerged as a powerful paradigm for solving tasks that require reasoning, planning, and interaction with external environments \citep{zhou2025agentfly, hu2025owl, zhang2025co, qiu2025alita, H2Oaigpte2025, 2025mirothinker, qiu2025alitag}. By combining language understanding with tool use, retrieval, and long-term memory, these agents show strong adaptability across a wide range of domains, from code generation to specialized scientific and scholarly domains \citep{yang2024sweagent, qiu2025agentdistill, li2025codetree, wang2025geneagent, qiu2025path, ding2025scitoolagent}. However, not every task requires the complex capabilities of agents, such as multi-step reasoning or long-context management (see Fig.~\ref{fig:motivation}). 
Contemporary LLMs, trained on web-scale corpora and in many cases coupled with web search tools (for example, GPT with online search in production APIs \citep{OpenRouterWebSearch2025}), can already solve a wide range of factual and well-structured queries with a single forward inference while with much lower computational and latency cost than a multi-step agent.

Hence, the central challenge now is to characterize the intelligence boundary of LLMs, enabling direct LLM inference within the boundary and escalating to agent only for tasks that exceed it. 
LLM query routing offers a practical way to probe this boundary by 
dynamically dispatching the query to models of varying quality and cost. 
Yet, existing research primarily focuses on routing exclusively among LLMs or among agents \citep{zhang2025router, zhang2025agentrouter, yue2025masrouter, liu2025rcr}, leaving the hybrid routing problem between LLMs and agents largely unexplored. 

\begin{figure}[t]
\centering
\includegraphics[width=0.95\textwidth]{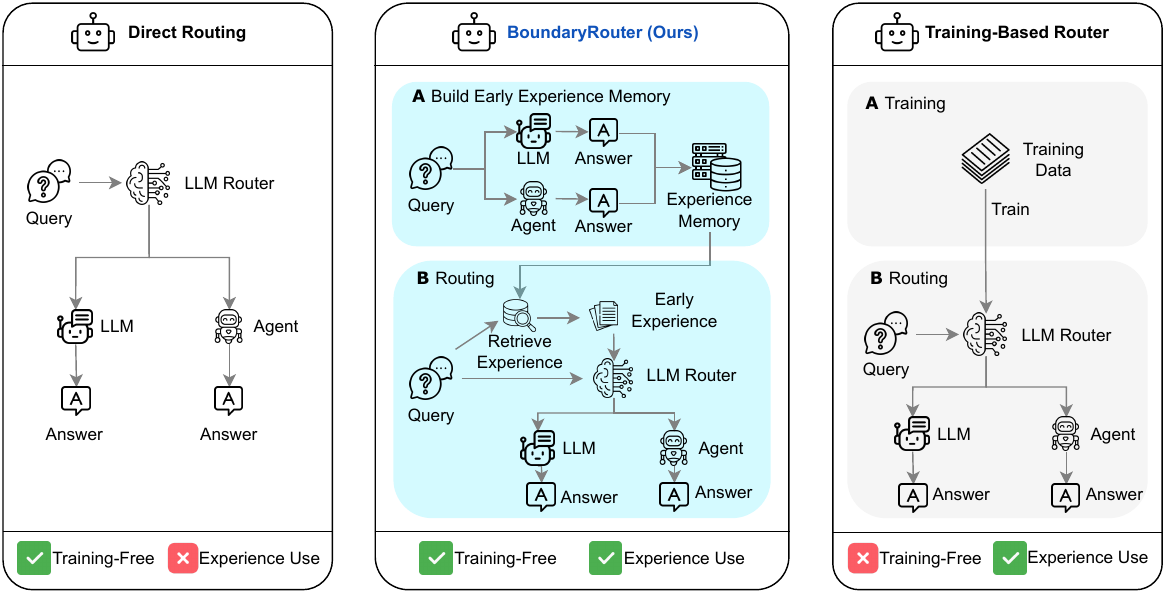}
    \caption{ Comparison of routers. \textit{Left}: Direct routing uses an LLM router to choose between direct LLM inference and full agent execution, but does not leverage experience. \textit{Right}: Training-based routing learns a router from labeled training data, enabling experience use but requiring supervision. \textit{Middle}: BoundaryRouter (ours) is training-free yet experience-driven: it first builds an early experience memory by running both the LLM and the agent on a small shared seed set, then retrieves similar experiences at test time to guide routing decisions.
    }
    \label{fig:routing_compare}
\end{figure}
To address this issue, we propose \textbf{BoundaryRouter}, a training-free query routing method that efficiently combines direct LLM inference with agentic execution 
through early experience and structured reasoning. 
A critical constraint in real-world routing is the cold-start problem: we often lack prior performance data (ground truth) for incoming queries and therefore cannot train a supervised router. BoundaryRouter addresses this by utilizing \textit{early experience}, a compact memory built by executing both the LLM and the agent on a shared seed set without knowing the ground truth, as shown in Figure \ref{fig:routing_compare}. 
Rather than serving as supervision or calibration data, this early experience acts as a lightweight behavioral reference that exposes systematic differences between the two systems.

To systematically study this routing problem, we construct \textbf{RouteBench}, a benchmark specifically designed to evaluate the decision boundaries of LLMs to route between LLMs and agents.
Unlike conventional evaluation suites that assume a static distribution, RouteBench assesses routing generalization across three progressively challenging dimensions: standard in-domain tasks, linguistically perturbed queries for robustness, and out-of-domain scenarios. This design enables a rigorous assessment of how well routers can balance performance and cost when facing both familiar and novel task distributions.

Finally, we evaluate BoundaryRouter on Routebench, which reduces average inference time by \textbf{60.6}\% compared to the agent and achieves \textbf{28.6}\% performance improvement over direct LLM inference, demonstrating a clearly better cost–performance trade-off.
Using BoundaryRouter, we further evaluate 14 contemporary models on RouteBench. Among frontier models, GPT-5, Gemini-3-Pro-Preview, and Gemini-2.5-Pro achieve the strongest overall routing performance. Compared with simple prompt routing or retrieval-augmented generation (RAG) routing, BoundaryRouter improves routing quality by \textbf{37.9}\% and \textbf{8.2}\%, respectively, confirming the effectiveness of early experience and rubric-guided reasoning.
Our findings highlight that in cold-start settings without routing labels or ground-truth, early behavioral signals paired with rubric-constrained reasoning can enable reliable routing between LLMs and agents, offering a practical path toward scalable coordination in heterogeneous reasoning systems.

\begin{figure*}[t]
    \centering

    \begin{subfigure}{0.46\textwidth}
        \centering
        \includegraphics[width=\linewidth]{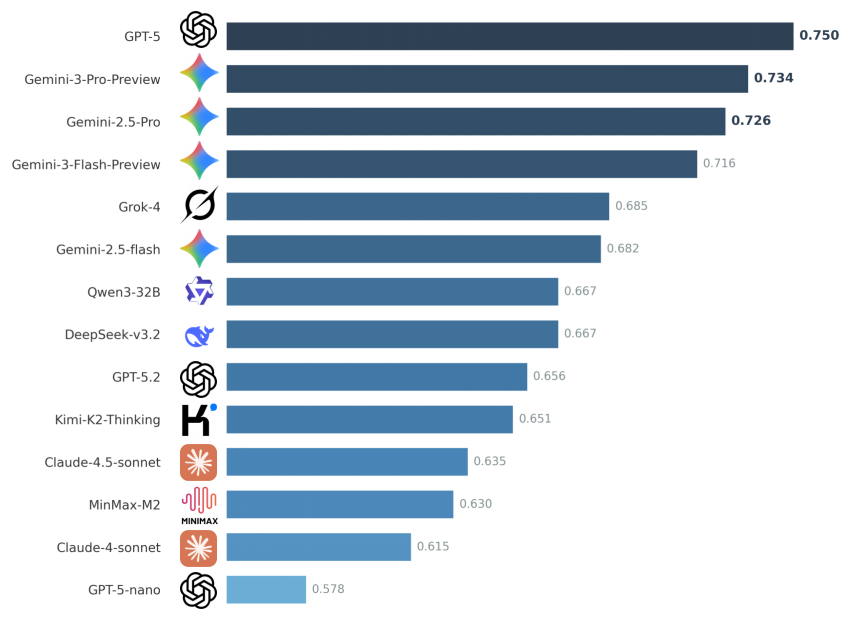}
        \caption{Average RouteBenchScore across models.}
        \label{fig:model-performance}
    \end{subfigure}
    \hfill
    \begin{subfigure}{0.52\textwidth}
        \centering
        \includegraphics[width=\linewidth]{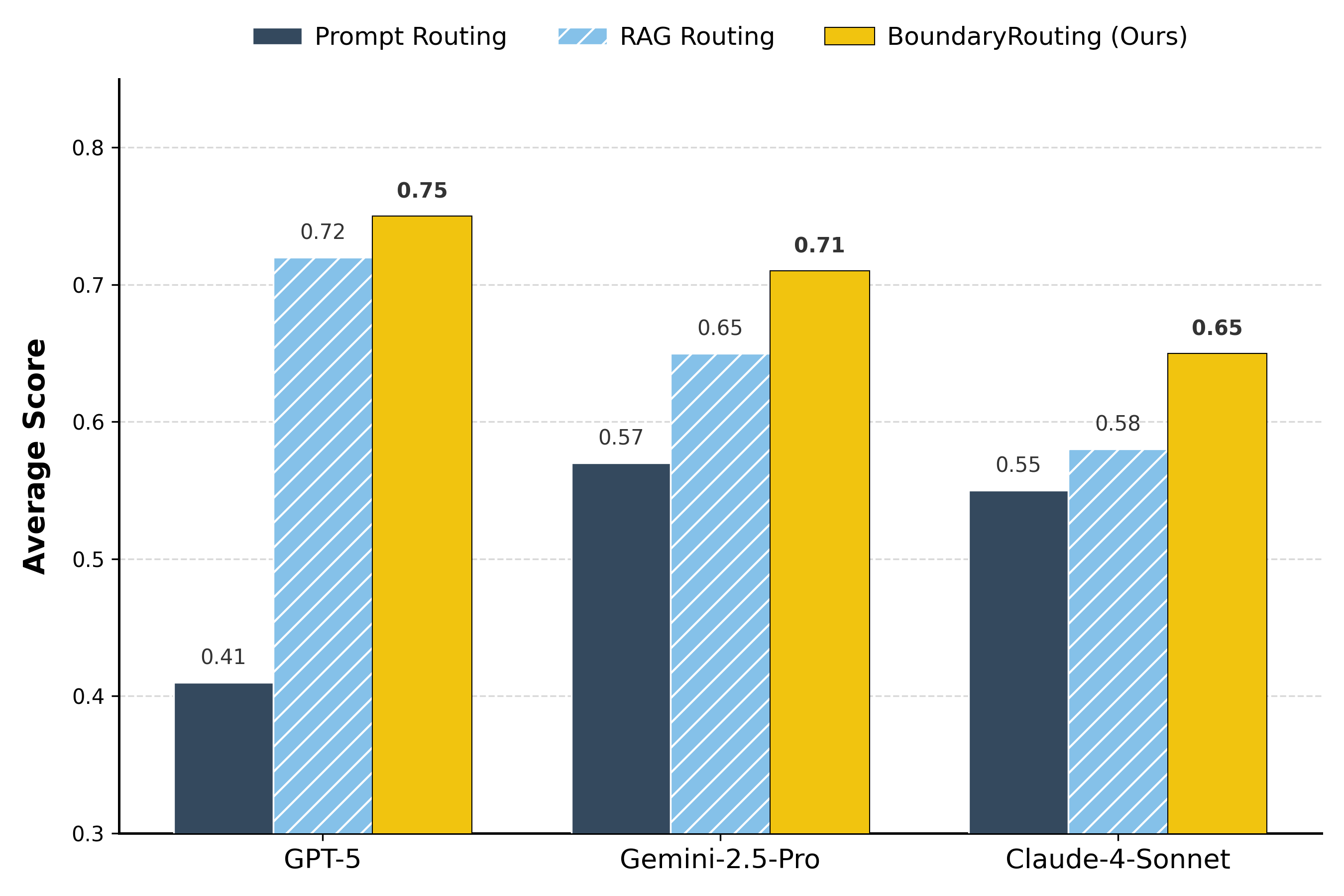}
        \caption{Overall Routing Effectiveness of BoundaryRouter.}
        \label{fig:ablation}
    \end{subfigure}

    \caption{\textbf{Overall routing performance and cost trade-offs on RouteBench}.(a) Average RouteBenchScore across all evaluation sets for different models, sorted in descending order; (b) Comparison of routing effectiveness across different routing strategies. The bar chart reports the average routing score on RouteBench for basic prompt-based routing, retrieval-based (RAG) routing, and our routing method across three backbone models. Our method consistently achieves higher routing performance than both baselines, demonstrating its effectiveness as a general routing strategy.}
    \label{fig:twofigs}
\end{figure*}


\section{Related Works}
\textbf{Routing in LLM, LLM Agents and Multi-Model Systems.} 
Task routing has become increasingly central to scaling language model systems, especially as workloads grow in diversity and cost sensitivity. Early efforts established the foundation by evaluating routing performance across diverse benchmark datasets \citep{shnitzer2023large}. A series of recent works explore how to dynamically dispatch queries across multiple models, agents, or tools to optimize utility through universal routing frameworks \citep{jitkrittum2025universal}. 
Some approaches, like Router-R1 \citep{zhang2025router}, integrate routing into multi-step inference, where models iteratively decide which component to consult based on intermediate reasoning signals, often using reinforcement learning to balance accuracy and cost. Others operate in multi-agent setups, coordinating agents with different roles or specializations via hierarchical planning, graph-based dispatching, or role-aware context filtering \citep{yue2025masrouter, zhang2025agentrouter, liu2025rcr}. Routing under resource constraints has also received attention, with methods selecting models adaptively based on utility-cost trade-offs \citep{panda2025adaptive}, lookahead mechanisms \citep{huang2025lookahead}, or test-time compute optimization \citep{ding2025best}. Furthermore, reward-guided ensembles have been proposed to route queries to the most capable expert model \citep{lu-etal-2024-routing}. In parallel, retrieval-augmented reasoning systems treat routing as a step-wise selection over knowledge bases or tools \citep{peng2025learning}. While these directions reflect growing interest in adaptive coordination, they typically focus on routing within homogeneous spaces, across models, agents, or tools, but not across them. Our work fills this gap by studying routing between LLMs and agents, enabling task-level decisions that exploit their complementary strengths.

\textbf{Learning from Early Experience and Self-Evolving Agents.} 
To build more adaptive and autonomous systems, recent work has explored how agents can learn from their early history. Reflexion \cite{shinn2023reflexion} and Voyager  \cite{wang2023voyager} demonstrate that agents can reflect on failures, consolidate long-term memory, and develop reusable skills through language-based feedback and exploration. Beyond retrospective improvement, some methods adapt agents at test time by detecting errors or uncertainty and updating internal components accordingly  \cite{acikgoz2025self}, while others leverage early interaction data to bootstrap policies via future-consistent behavior modeling \cite{zhang2025agent}.
These ideas have also reshaped how agents are architected. Instead of relying on static pipelines, agents can evolve dynamically by refining their internal logic, memory, and prompting strategies over time \cite{wu2025evolver}. A recent survey consolidates these directions into the emerging framework of self-evolving agents, which emphasizes the shift from static models to continually adapting, self-refining systems \citep{gao2025survey}.
We extend experience-based learning from task execution to routing, enabling agents to improve delegation decisions across LLMs and agents, a dimension rarely addressed in prior work.

\textbf{Reasoning-Enhanced Decision Making.} Chain-of-Thought (CoT) prompting, introduced by Wei et al. \citep{wei2022chain}, enables large language models to reason through intermediate steps before producing final answers. Beyond accuracy gains, CoT supports decision-making within models and agents. In ReAct \citep{yao2022react}, reasoning traces guide tool use and subroutine selection, while other work employs CoT for task decomposition and option evaluation \citep{zhou2022least, kojima2022large}.
Recent advances incorporate rubric-guided CoT, where reasoning is shaped by explicit evaluation criteria rather than implicit preferences. This approach improves consistency and alignment in tasks such as text generation, code evaluation, and geospatial planning by ensuring that reasoning respects domain-specific constraints \citep{pathak2025rubric, chen2025empowering}.
Building on this direction, we apply CoT to agent routing, helping models reason explicitly, under rubric guidance, about which agent or submodel is best suited for a given task. This reframes CoT as a structured coordination mechanism that enhances routing transparency and reliability.

\begin{figure*}[t]
    \centering
    \includegraphics[width=0.95\textwidth]{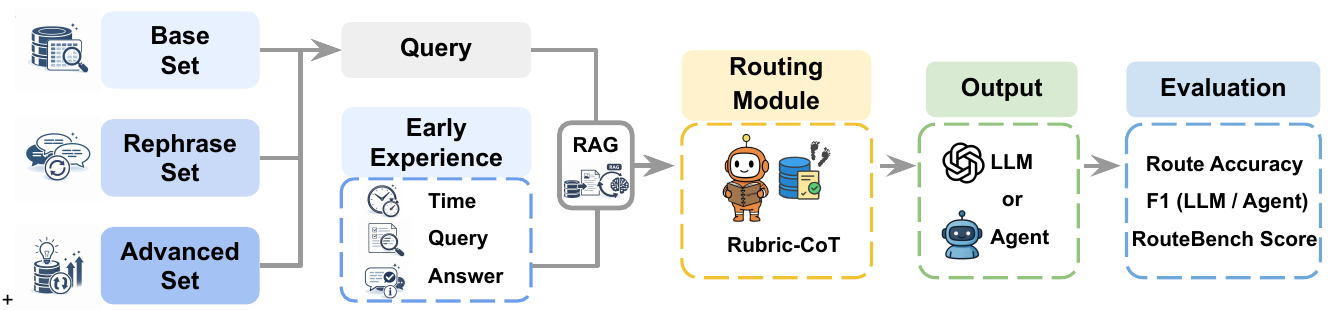}
    \caption{Overview of our routing pipeline. A query from RouteBench is routed using early-experience retrieval-augmented (RAG) and a CoT-based routing module, which delegates the task to either an LLM or an Agent. Performance is evaluated on RouteBench via routing accuracy and per-solver F1, aggregated into a RouteBenchScore.
}
    \label{fig:pipeline}
\end{figure*}

\section{Learning Agent Routing from Early Experience}

\subsection{Routing Module Overview}
The routing module is an LLM-based decision system that routes each incoming query to either a lightweight LLM (low latency and token cost) or a full agent (higher cost but stronger tool-augmented reasoning). Concretely, the router is a \emph{pluggable} routing LLM that conditions on (i) the input query and (ii) retrieved early-experience cases containing solver outputs and runtime.
Given these signals, the router follows a rubric-guided reasoning (Box~\ref{rules}) to trade off expected answer quality against latency and then outputs the routing decision.
Formally, given an input query $x$, the router produces $\text{Route}(x) \in \{\text{LLM},\ \text{Agent}\}$. 
Importantly, the routing module is \textbf{training-free}: it uses no gradient updates and does not require supervision on routing labels. This design makes the router easy to deploy and to adapt to new tasks or new agent implementations by updating only the early-experience memory and retrieval components.



\subsection{Learning from Early Experience}
\label{subsec:early_experience}

A central difficulty in LLM--agent routing is \emph{cold start}: when the router is deployed, we typically do not have ground-truth for the incoming queries and therefore cannot obtain reliable routing labels. This makes standard supervised router training impractical. To address this, we introduce \textbf{Learning from Early Experience}, which provides the router with a compact memory of observable solver behavior, with the pipeline overview shown in Fig.~\ref{fig:pipeline}.

\textbf{Constructing the early-experience memory.}
We first sample a small seed set of questions \(\mathcal{D}_{\mathrm{seed}}\) and run both candidate solvers---a lightweight LLM and a full agent---on the same inputs. For each \(x \in \mathcal{D}_{\mathrm{seed}}\), we record only deployment-time observable information:
\begin{itemize}
    \item the question \(x\),
    \item the LLM output \(y^{\mathrm{LLM}}\) and latency \(t^{\mathrm{LLM}}\),
    \item the agent output \(y^{\mathrm{Agent}}\) and latency \(t^{\mathrm{Agent}}\).
\end{itemize}
Crucially, we do not store gold answers, correctness labels, or rewards. The resulting memory
\[
\mathcal{M} = \{(x_i, y_i^{\mathrm{LLM}}, y_i^{\mathrm{Agent}}, t_i^{\mathrm{LLM}}, t_i^{\mathrm{Agent}})\}_{i=1}^{N}
\]
captures systematic differences in the two systems' behavior (e.g., response and runtime) without requiring supervision.


\textbf{Retrieval-augmented routing.}
At inference time, given a new query \(x\), we retrieve the top-\(K\) most similar records from \(\mathcal{M}\) using a hybrid retriever (sparse lexical matching plus dense semantic similarity):
\[
\text{Retrieve}(\mathcal{M}, x) = \{(x_k, y_k^{\mathrm{LLM}}, y_k^{\mathrm{Agent}}, t_k^{\mathrm{LLM}}, t_k^{\mathrm{Agent}})\}_{k=1}^{K}.
\]
These retrieved cases are provided to the routing LLM as evidence. The router compares the current query against the retrieved questions and inspects the two solvers' outputs and latencies to infer regularities, e.g., whether similar questions previously led the agent to produce slower, multi-step reasoning, or how similarity in phrasing or structure correlates with the relative efficiency and behavior of the two systems.

\subsection{Rubric-guided Chain-of-Thought Routing}
\label{subsec:rubric_guided_cot}
In the cold-start setting, the router must make a decision without access to ground-truth answers or routing labels.
A natural approach to routing is to allow the routing LLM to reason explicitly before selecting between the LLM and the agent. 
Chain-of-Thought (CoT) prompting has been shown to improve structured decision-making by encouraging step-by-step reasoning 
\citep{wei2022chain, yao2022react, kojima2022large}. 
However, in our setting, routing is not an open-ended reasoning task: decisions should follow explicit behavioral criteria, 
such as comparing answer characteristics and response times observed in early experience, as formalized in 
Box~\ref{rules}. 

Direct, free-form CoT does not guarantee that the routing LLM will consistently attend to these criteria, especially under 
paraphrasing or distribution shift. To align the reasoning process with the rule-based nature of routing, we therefore adopt 
a rubric-guided CoT formulation that explicitly encodes the evaluation protocol into the prompt. 
As shown in Fig.~\ref{fig:Prompt Comparison}, the router is required to follow a fixed decision rubric that reflects the 
actual dimensions available in the early-experience memory, rather than relying on unconstrained reasoning.

\begin{figure*}[t]
    \centering
    \includegraphics[width=0.95\textwidth]{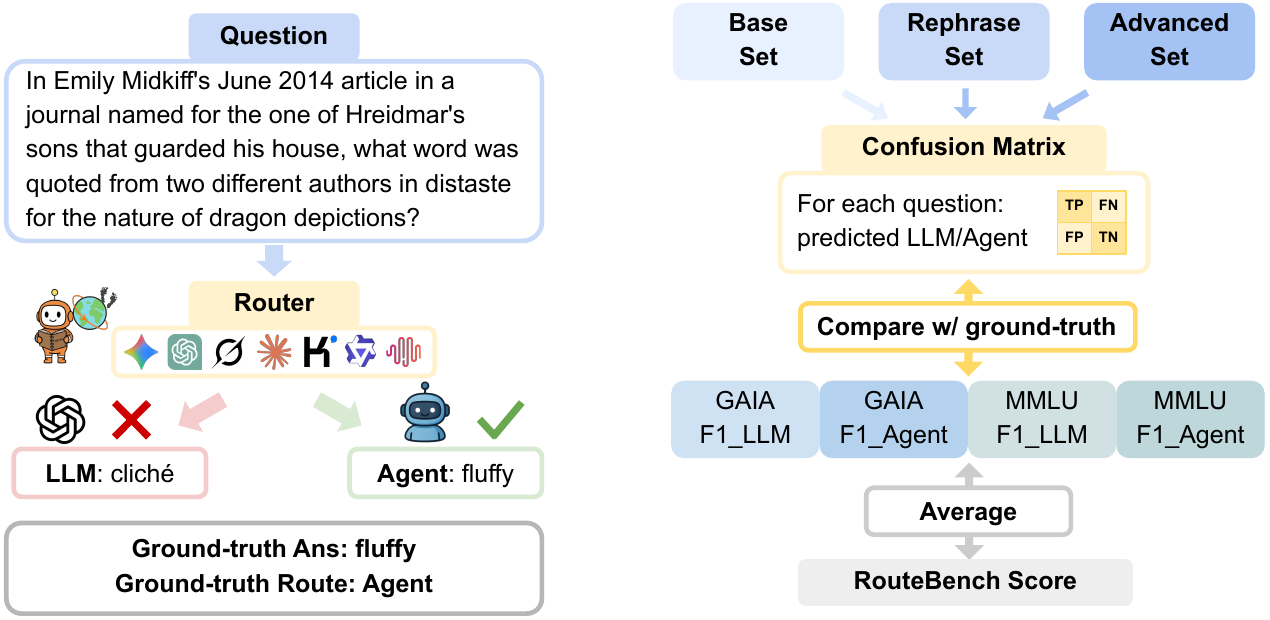}
    \caption{Overview of the RouteBench evaluation framework. On the left is the Single-instance routing process: solver outputs determine the ground-truth route, and the router’s decision is evaluated via exact-match accuracy. On the right is Set-level evaluation: the router’s predictions over all questions yield per-solver F1 scores (LLM vs. Agent), which are averaged to produce the final RouteBenchScore.}
    \label{fig:routebench}
\end{figure*}
\section{RouteBench: Benchmarking LLM Routing between LLM and Agent}




Routing between heterogeneous reasoning systems requires a benchmark with diverse tasks, clear supervision, and controlled distribution shifts.
To address this, we introduce \textbf{RouteBench}, a benchmark designed to evaluate how effectively a model assigns queries between a lightweight LLM and a full agent.
RouteBench consists of a curated question pool drawn from GAIA and MMLU, paired solver outputs from both systems, and human-annotated routing labels.
Each instance includes the question, solver predictions, latency, and the ground-truth routing decision.
To assess routing generalization, RouteBench provides three evaluation sets covering in-domain, paraphrased, and out-of-domain conditions.
Routing performance is evaluated at both the instance level and the set level, using routing accuracy, solver-specific F1, and the final RouteBenchScore (Fig.~\ref{fig:routebench}).

\subsection{Benchmark Curation}
To form the base question pool, we sample from two established sources. GAIA~\citep{mialon2023gaia} provides open-ended reasoning tasks that reflect real-world problem solving. MMLU~\citep{hendryckstest2021} spans 57 academic subjects and provides structured knowledge questions. We randomly sample \textbf{30 GAIA questions} and \textbf{57 MMLU questions} (one per subject), producing a compact yet diverse collection that covers factual recall, symbolic reasoning, multi-step planning, and open-domain inference.

For each selected question, we collect solver predictions from two systems: an LLM and a full agent with tool-use and intermediate reasoning capabilities. All samples undergo manual review to ensure semantic clarity, correctness of solver traces, and consistency of formatting. This curated pool serves as the foundation for all evaluation splits introduced in later sections.

\subsection{Benchmark Composition}
Each RouteBench instance can be viewed as a 5-tuple
$$(x, y^{\mathrm{LLM}}, y^{\mathrm{Agent}}, y^{*}, d), $$
where these elements correspond to the question, LLM prediction,
agent prediction, ground-truth answer, and routing decision, respectively.
The first four fields describe the task and solver behaviors. The final field, $d$, is the human-annotated label indicating which solver provides the preferred answer for that question.

The label of $d$ follows a fixed deterministic rule:

\begin{tcolorbox}[
  colback=gray!5!white,
  colframe=gray!50!black,
  title=Ground-truth Routing Rule,
  fonttitle=\bfseries,
  breakable,
  enhanced,
  sharp corners=south,
  colbacktitle=gray!10!black,
  label={rules},
  before upper=\refstepcounter{equation}
]
\small 
\textbf{1. Correctness priority:}  
If only one solver produces the correct answer, choose that solver.

\textbf{2. Efficiency tie-break:}  
If both solvers are correct, choose the one with shorter response time.

\textbf{3. Failure fallback:}  
If both solvers are incorrect, choose the agent, which has a higher chance of recovery through multi-step reasoning.
\end{tcolorbox}

This labeling scheme ensures that RouteBench measures the ability to infer better solver selection, rather than answer-generation accuracy.
\subsection{Evaluation Sets for Routing Generalization}

From the curated question pool, we construct three evaluation sets designed to assess routing under different generalization conditions.

\textbf{Base Set (In-domain).}
The Base Set contains 30 GAIA and 57 MMLU questions with solver predictions and routing labels. It reflects the in-domain distribution and also serves as the early-experience corpus for retrieval-augmented routing. All questions are evaluated in their original form.

\textbf{Rephrase Set (Paraphrased In-domain).}
The Rephrase Set is created by paraphrasing each Base Set question using a controlled LLM-based rewriting process that preserves semantics while modifying surface form. This set evaluates routing stability under linguistic variation without changing task content.

\textbf{Advanced Set (Out-of-domain).}
It consists of GAIA and MMLU questions that are disjoint from the Base and Rephrase Sets. Although drawn from the same benchmarks, these questions differ in topic and reasoning structure, providing a controlled out-of-domain evaluation.


\subsection{Evaluation Metrics}
\label{Evaluation Metrics}
RouteBench uses a small set of complementary metrics to evaluate routing and provide a single scalar score for model comparison. Instance-level accuracy and solver-level PRF metrics serve as diagnostic measures, while RouteBenchScore is the primary metric for overall routing quality.

\textbf{Instance-level routing accuracy.}
For each question, the routing model outputs a binary decision indicating which solver to use. We compute exact-match accuracy by comparing predictions against the human-annotated ground-truth routing labels defined in~\ref{rules}. This metric reflects overall decision correctness but does not provide detailed information about each solver’s routing characteristics.

\textbf{Solver-level PRF metrics.}
To characterize routing behavior for each solver, we compute Precision, Recall, and F1 for the LLM and the agent separately.
Let class~A denote routing to the LLM and class~B denote routing to the agent.
For each class, RouteBench provides the number of ground-truth assignments,
$\text{tot}_A$ and $\text{tot}_B$, while a routing model yields true positives,
$\text{tp}_A$ and $\text{tp}_B$.
False positives and false negatives follow by symmetry,
$\text{fp}_A = \text{fn}_B$ and $\text{fp}_B = \text{fn}_A$.
Precision, Recall, and F1 are then computed independently for each solver within each evaluation set.

\textbf{Final RouteBench score.}
To obtain a single summary metric, we average the solver-level F1 scores across both solvers and task sources:
\[
\text{RouteBenchScore}
= \frac{1}{4} \sum_{s} \sum_{d} \text{F1}_{s}^{d}.
\]
Here, $s$ ranges over the two solvers (LLM and agent), and $d$ ranges over the two task sources (GAIA and MMLU).
This score reflects how consistently a routing model selects the correct solver across both open-domain and academic reasoning tasks, while weighting the two solvers and the two benchmarks equally.


\textbf{Primary comparison metric.}
Although instance-level accuracy and solver-level PRF metrics are useful for diagnostic analysis, \textbf{RouteBenchScore is the primary metric used for comparing routing models}. All model-to-model comparisons in this paper are based on this score, as it provides a concise summary of overall routing effectiveness across evaluation settings.

\section{Experiment}
\subsection{Experiment Setup}
\textbf{Models.}
To construct the early-experience database used for retrieval-augmented routing, we adopt GPT-4o~\cite{hurst2024gpt} as the representative LLM.
For the agent implementation, we use Claude Sonnet~4 for agent logic and GPT-4o as the tool-calling backbone within the SmolAgent Open DeepResearch framework~\cite{smolagents}.
We evaluate routing performance across a broad set of contemporary large language models, covering both state-of-the-art systems and widely deployed alternatives.
The evaluated models include GPT-5~\cite{openai2025gpt5}, GPT-5.2~\cite{openai2025gpt52}, GPT-5-nano~\cite{openai2025gpt5}, Gemini-3-Pro-Preview~\cite{deepmind_gemini3pro}, Gemini-3-Flash-Preview~\cite{google2025gemini3flash}, Gemini-2.5-Pro~\cite{comanici2025gemini}, Gemini-2.5-flash~\cite{comanici2025gemini}, Claude Sonnet 4, Claude Sonnet 4.5~\cite{anthropic2025sonnet45}, MinMax-M2~\cite{minimax2025m2},	Qwen3-32b~\cite{yang2025qwen3}, Grok-4~\cite{xai2025grok4}, Kimi-K2-Thinking~\cite{moonshotai2025kimik2thinking}, DeepSeek-v3.2~\cite{liu2025deepseek}. 

\textbf{Evaluation Details.}
All experiments are conducted on RouteBench.
For each question, the resulting decision is compared against the ground-truth routing decision, $d$.
Performance is computed using the official RouteBench scoring procedure described in Section~\ref{Evaluation Metrics}. In all experiments, the same BoundaryRouter framework is used, with different LLMs adopted as the routing model.
All API-based models are evaluated using the default parameters provided by Openrouter.

\begin{table*}[t]
\centering
\caption{\textbf{Routing performance and inference cost on RouteBench}.
Results are reported for LLM-only, Agent-only, and our routing method on the Base, Rephrase, and Advanced evaluation sets.
For each set, we report routing accuracy (Acc.) and average inference time (Time, in seconds) on MMLU, GAIA, and their average (Avg).
Our routing method consistently achieves higher accuracy than the LLM-only baseline while substantially reducing inference cost compared to the Agent-only baseline across all settings, striking a balance between the performance and cost.
}
\small
\resizebox{\textwidth}{!}{
\setlength{\tabcolsep}{3.5pt}
\begin{tabular}{l|
cccccc |
cccccc |
cccccc}
\toprule
& \multicolumn{6}{c}{\textbf{Base Set}}
& \multicolumn{6}{c}{\textbf{Rephrase Set}}
& \multicolumn{6}{c}{\textbf{Advanced Set}} \\
\textbf{Model}
& \multicolumn{2}{c}{MMLU} & \multicolumn{2}{c}{GAIA} & \multicolumn{2}{c}{Avg}
& \multicolumn{2}{c}{MMLU} & \multicolumn{2}{c}{GAIA} & \multicolumn{2}{c}{Avg}
& \multicolumn{2}{c}{MMLU} & \multicolumn{2}{c}{GAIA} & \multicolumn{2}{c}{Avg} \\
& Acc. & Time & Acc. & Time & Acc. & Time
& Acc. & Time & Acc. & Time & Acc. & Time
& Acc. & Time & Acc. & Time & Acc. & Time \\
\midrule
LLM                 & 0.754 & 2.22 & 0.1 & 8.633 & 0.528 & 4.431  & 0.754 & 0.533 & 0.13 & 5.58 & 0.54 & 2.27  & 0.860 & 4.17 & 0.23 & 5.23 & 0.64 & 4.53 \\
Agent               & 0.895 & 174.67 & 0.533 & 436.60 & 0.77 & 264.99  & 0.895 & 112.16 & 0.6 & 279.80 & 0.793 & 169.70  & 0.982 & 147.32 & 0.667 & 393.428 & 0.87 & 232.18 \\
\midrule
BoundaryRouter      & 0.8772 & 26.6 & 0.4 & 244.86 & 0.713 & 101.86  & 0.842 & 24.35 & 0.467 & 222.89 & 0.713 & 92.81  & 0.877 & 36.08 & 0.567 & 197.03 & 0.77 & 91.58 \\
\bottomrule
\end{tabular}
}

\label{tab:minitest_results}
\end{table*}
\subsection{Main Results}
Table~\ref{tab:minitest_results} summarizes the accuracy-latency trade-off on the three splits (Base, Rephrase, Advanced), reported separately on MMLU and GAIA, and averaged across two sources.

\textbf{Performance of BoundaryRouter.}
Across all three evaluation sets, BoundaryRouter consistently achieves a favorable balance between accuracy and inference cost.
Compared to the LLM-only run, BoundaryRouter substantially improves accuracy on both MMLU and GAIA, with an average relative improvement of 28.6\%.
Compared to Agent-only routing, our method reduces inference time by an order of magnitude, with a \textbf{60.6}\% relative reduction, while retaining a large fraction of the agent’s accuracy, only a relative 11.5\% decrease.
On the Base Set, the LLM-only baseline exhibits low GAIA accuracy (0.10) despite fast inference, whereas the Agent achieves higher accuracy at a prohibitive average cost of 264.99 seconds. Our routing method bridges this gap, improving average accuracy to 0.713 while reducing inference time to 101.86 seconds.
A similar trend is observed on the Rephrase Set. 
Crucially, in the out-of-domain set, the Advanced Set, BoundaryRouter still performs well. It achieves a strong accuracy of 0.77 with less than half of the Agent’s inference time. This result indicates that the routing strategy generalizes beyond the in-domain distribution and remains effective under distribution shift.
These results show that BoundaryRouter effectively routes easy questions to the LLM and leaves hard questions for the Agent.
This selective invocation allows the system to maintain strong performance while avoiding the high cost of invoking the agent for every query.

\textbf{Comparison with LLM-only and Agent-only baselines.}
Overall, while the Agent achieves 43.7\% higher accuracy than the LLM baseline, it is nearly \textbf{60}$\times$ slower in inference time. Neither baseline alone provides a satisfactory balance between accuracy and inference cost.
The LLM baseline offers low latency but limited performance on GAIA, while the Agent baseline improves performance at the cost of extremely high inference time. BoundaryRouter bridges this gap by combining the strengths of both approaches, achieving strong accuracy while maintaining a much lower inference cost.


\begin{table}[tb]
  \centering
  \caption{\textbf{Model Performance on RouteBench}. Models are ranked from top to bottom by their overall average score. All rankings are computed using the full-precision underlying scores, while values reported in the table are rounded to two decimal places for readability, but rounded to three decimal places for the overall average for easier comparison.}
  \resizebox{\textwidth}{!}{%
    \setlength{\aboverulesep}{0pt}
    \setlength{\belowrulesep}{0pt}
    \renewcommand{\arraystretch}{1.1}
    
    \begin{tabular}{l c rrrr r rrrr r rrrr r}
    \toprule
    
    \multirow{3}{*}{Model} & \multirow{3}{*}{Avg.} & \multicolumn{5}{c}{Base Set} & \multicolumn{5}{c}{Rephrase Set} & \multicolumn{5}{c}{Advanced Set} \\
    \cmidrule(lr){3-7} \cmidrule(lr){8-12} \cmidrule(lr){13-17}
    
    & & \multicolumn{2}{c}{MMLU} & \multicolumn{2}{c}{GAIA} & \multirow{2}{*}{Avg.} & \multicolumn{2}{c}{MMLU} & \multicolumn{2}{c}{GAIA} & \multirow{2}{*}{Avg.} & \multicolumn{2}{c}{MMLU} & \multicolumn{2}{c}{GAIA} & \multirow{2}{*}{Avg.} \\
    \cmidrule(lr){3-4} \cmidrule(lr){5-6} \cmidrule(lr){8-9} \cmidrule(lr){10-11} \cmidrule(lr){13-14} \cmidrule(lr){15-16}
    
    & & LLM & Agent & LLM & Agent & & LLM & Agent & LLM & Agent & & LLM & Agent & LLM & Agent & \\
    \midrule
    
    GPT-5             & \cellcolor{gray!20}0.750 & 0.95 & 0.86 & 0.60 & 0.92 & \underline{0.83} & 0.95 & 0.86 & 0.55 & 0.90 & \textbf{0.81} & 0.81 & 0.32 & 0.46 & 0.85 & \underline{0.61} \\
    Gemini-3-Pro-Preview             & \cellcolor{gray!20}0.734 & 0.94 & 0.83 & 0.55 & 0.90 & 0.80 & 0.95 & 0.86 & 0.44 & 0.87 & \underline{0.78} & 0.79 & 0.24 & 0.55 & 0.90 & \textbf{0.62} \\
    Gemini-2.5-Pro   & \cellcolor{gray!20}0.726 & 0.93 & 0.80 & 0.67 & 0.94 & \textbf{0.84} & 0.93 & 0.80 & 0.36 & 0.86 & 0.75 & 0.70 & 0.38 & 0.44 & 0.90 & \underline{0.61} \\
    Gemini-3-Flash-Preview   & \cellcolor{gray!20}0.716 & 0.95 & 0.86 & 0.50 & 0.88 & 0.80 &  0.94 & 0.83 & 0.44 & 0.87 & 0.77 & 0.76 & 0.22 & 0.44 & 0.90 & 0.58 \\
    Grok-4            & \cellcolor{gray!20}0.685 & 0.95 & 0.86 & 0.44 & 0.90 & 0.79 & 0.95 & 0.86 & 0.43 & 0.83 & 0.77 & 0.73 & 0.31 & 0.17 & 0.79 & 0.50 \\
    Gemini-2.5-flash  & \cellcolor{gray!20}0.682 & 0.95 & 0.86 & 0.50 & 0.88 & 0.80 & 0.93 & 0.80 & 0.36 & 0.86 & 0.74 & 0.74 & 0.15 & 0.33 & 0.83 & 0.51 \\
    Qwen3-32b         & \cellcolor{gray!20}0.667 & 0.97 & 0.89 & 0.35 & 0.74 & 0.74 & 0.95 & 0.85 & 0.27 & 0.76 & 0.71 & 0.81 & 0.26 & 0.42 & 0.73 & 0.56 \\
    DeepSeek-v3.2         & \cellcolor{gray!20}0.667 & 0.92 & 0.76 & 0.38 & 0.78 & 0.71 & 0.94 & 0.83 & 0.46 & 0.80 & 0.76 &  0.84 & 0.29 & 0.27 & 0.76 & 0.54 \\
    GPT-5.2         & \cellcolor{gray!20}0.656 & 0.93 & 0.79 & 0.38 & 0.77 & 0.72 & 0.94 & 0.80 & 0.46 & 0.80 & 0.75 & 0.76 & 0.08 & 0.40 & 0.80 & 0.51 \\
    Kimi-K2-Thinking  & \cellcolor{gray!20}0.651 & 0.93 & 0.77 & 0.40 & 0.80 & 0.73 & 0.92 & 0.76 & 0.44 & 0.76 & 0.72 & 0.80 & 0.18 & 0.33 & 0.71 & 0.51 \\
    Claude-4.5-sonnet & \cellcolor{gray!20}0.635 & 0.95 & 0.85 & 0.33 & 0.71 & 0.71 &  0.95 & 0.85 & 0.68 & 0.32 & 0.70 & 0.85 & 0.22 & 0.29 & 0.62 & 0.49 \\
    MinMax-M2   & \cellcolor{gray!20}0.630 & 0.86 & 0.60 & 0.35 & 0.74 & 0.64 & 0.90 & 0.73 & 0.25 & 0.73 & 0.65 & 0.75 & 0.32 & 0.50 & 0.82 & 0.60 \\
    Claude-4-sonnet   & \cellcolor{gray!20}0.615 & 0.93 & 0.77 & 0.27 & 0.76 & 0.68 &  0.95 & 0.78 & 0.33 & 0.71 & 0.69 & 0.84 & 0.00 & 0.33 & 0.71 & 0.47 \\
    GPT-5-nano        & \cellcolor{gray!20}0.578 & 0.85 & 0.63 & 0.35 & 0.74 & 0.64 &  0.87 & 0.65 & 0.22 & 0.66 & 0.60 & 0.78 & 0.34 & 0.32 & 0.51 & 0.49 \\
    
    \bottomrule
    \end{tabular}
  } 
 \label{tab:model_performance_hierarchical}
\end{table}
\subsection{RouteBench Results}
\textbf{Overall routing performance}.
Table~\ref{tab:model_performance_hierarchical} reports routing performance across 14 models on RouteBench. Models are ranked by their overall average score. We observe a clear performance stratification across model families. \textbf{GPT-5} achieves the strongest overall routing performance, with an average score of 0.75, followed closely by \textbf{Gemini-3-Pro-Preview} (0.734) and \textbf{Gemini-2.5-Pro} (0.726). These models remain stable even when questions are rewritten or shifted to new topics, suggesting that their routing patterns are less sensitive to changes in surface form or content. These top-ranked models consistently exhibit strong solver discrimination on both MMLU and GAIA, indicating robust routing decisions across domains and distribution shifts. This finding also aligns with these models' abilities on other benchmarks, like AIME25~\footnote{\url{https://huggingface.co/datasets/yentinglin/aime_2025}}, LiveCodeBench~\cite{jain2024livecodebench}, and Humanity's last exam~\cite{phan2025humanity}.

\textbf{Routing is surprisingly still a relatively hard problem}. Even in the in-domain setting where questions closely match the early experience and the task is binary, routing accuracy remains far from perfect across all models. This difficulty becomes more pronounced under distribution shift, but its presence in the Base set already indicates that effective routing is not a solved problem, even without paraphrasing or topic change.

\textbf{Advanced Set Drives Ranking Separation}. While routing performance is similar across models on the Base and Rephrase sets, differences become large on the Advanced set, where top models retain average scores around 0.61–0.62, mid-tier models fall to 0.54–0.56, and lower-ranked models drop below 0.50. This ranking reshuffle suggests that out-of-domain routing ability, rather than in-domain decision matching, is the primary factor distinguishing strong routing models from weaker ones.



\subsection{Ablation Study}
\begin{table*}[t]
\centering
\setlength{\tabcolsep}{3pt}
\renewcommand{\arraystretch}{1.05}
\caption{Comparison of the three routing variants across GPT-5, Gemini 2.5 Pro, and Claude Sonnet 4 on Base, Rephrase, and Advanced Set. The table shows that early-experience memory and structured reasoning together provide the strongest and most stable performance.}
\label{tab:ablation}
\resizebox{\textwidth}{!}{%
\begin{tabular}{l cccc cccc cccc}
\toprule
& \multicolumn{4}{c}{\textbf{GPT-5}}
& \multicolumn{4}{c}{\textbf{Gemini-2.5-Pro}}
& \multicolumn{4}{c}{\textbf{Claude-4-Sonnet}} \\
\cmidrule(lr){2-5} \cmidrule(lr){6-9} \cmidrule(lr){10-13}
& Base & Rephrase & Advanced & Avg.
& Base & Rephrase & Advanced & Avg.
& Base & Rephrase & Advanced & Avg. \\
\midrule
Prompt Routing & 0.48 & 0.48 & 0.26 & 0.41 & 0.55 & 0.57 & 0.58 & 0.57 & 0.52 & 0.57 & 0.56 & 0.55 \\
RAG Routing    & 0.80 & 0.80 & 0.54 & 0.72 & 0.76 & 0.77 & 0.42 & 0.65 & 0.61 & 0.60 & 0.51 & 0.58 \\
\midrule
BoundaryRouter & 0.83 & 0.81 & 0.61 & 0.75 & 0.78 & 0.79 & 0.54 & 0.71 & 0.73 & 0.65 & 0.56 & 0.65 \\
\bottomrule
\end{tabular}%
}
\end{table*}
To understand the contribution of each component in our routing framework, Table~\ref{tab:ablation} compares three variants: basic Prompt Routing, RAG Routing, and our Rucirc-guided CoT Routing with early experience (i.e., RAG), across GPT-5, Gemini 2.5 Pro, and Claude Sonnet 4. 
The corresponding prompts for the two ablation baselines are provided in Appendix~\ref{app: ablation_prompt}.

\textbf{Prompt Routing.}
This variant removes early-experience memory entirely. It selects between the LLM and the agent using only high-level capability profiles provided in the prompt, refer to Appendix~\ref{Prompt Routing}.
Without early experience, routing decisions depend strongly on surface cues, leading to unstable behavior under paraphrasing and distribution shift. 
Even in the Base set, average scores remain low (e.g., 0.41 for GPT-5 and 0.55 for Claude-4-Sonnet), indicating that capability-aware reasoning is insufficient for reliable solver selection.

\textbf{RAG Routing.}
This variant introduces early-experience memory but removes rubric-guided reasoning. Retrieved behavioral examples are shown to the router, which must directly output a binary routing decision without reasoning
(prompt in Appendix~\ref{app: ablation_prompt}).
It substantially improves performance, with an average of 27.5\% improvement, especially on the Base and Rephrase sets by introducing early-experience retrieval, suggesting that access to historical solver behavior provides useful signals for routing, even without the gold answer.


\textbf{Rubric-Guided CoT with Early Experience (BoundaryRouter).}
It combines early-experience memory with rubric-guided chain-of-thought reasoning, consistently achieves the best performance across all models, and sets, with an increase of 37.9\% compared to Prompt Routing  and of 8.2\% compared to RAG Routing.
Notably, it yields the highest Advanced-set scores for all three LLMs, also maintaining strong in-domain performance. This demonstrates that structured reasoning is essential for effectively interpreting retrieved experiences and making stable routing decisions under the distribution shift.

\section{Discussion and Conclusion}
We study cold-start routing between direct lightweight LLM inference and full agentic execution, where ground-truth labels are unavailable at deployment time. To address this, we propose BoundaryRouter, a training-free router that learns from early experience, and introduce RouteBench, a benchmark for evaluating LLM–agent routing under in-domain, paraphrased, and out-of-domain settings. Experiments show that BoundaryRouter improves the accuracy–latency trade-off over baselines and remains robust under paraphrasing and distribution shift. These results suggest that early experience provides useful signals for routing decisions even without access to ground-truth answers, and that structured reasoning helps maintain stable decisions when task distributions change. While our current framework focuses on binary routing between an LLM and a single agent pipeline, future work may explore more complex routing scenarios involving multiple agents or heterogeneous tools. Overall, our findings highlight routing as an important component for improving the efficiency and scalability of hybrid LLM–agent systems.
\newpage
\section*{Reproducibility Statement}

We provide sufficient details to enable full reproduction of RouteBench and the BoundaryRouter framework. RouteBench is constructed from publicly available GAIA and MMLU benchmarks with explicitly defined sampling, annotation, and evaluation procedures. Each instance includes solver outputs, latency, and deterministic routing labels based on a fixed decision rule. BoundaryRouter is training-free and relies on early-experience memory, retrieval, and rubric-guided routing, all of which are fully specified in the paper and appendix, including prompts and decision protocols. All models are evaluated using the same splits and scoring procedure, including routing accuracy and RouteBenchScore. We will release the RouteBench dataset, routing prompts, and evaluation code upon publication to ensure full reproducibility.

\bibliography{iclr2026_conference}
\bibliographystyle{iclr2026_conference}
\newpage
\appendix

\section{Appendix}
\subsection{Use of LLMs}
LLMs were used solely to improve the clarity and readability of the manuscript. 

\subsection{Example}
\begin{figure}[htbp]
\centering
    \includegraphics[width=0.95\textwidth]{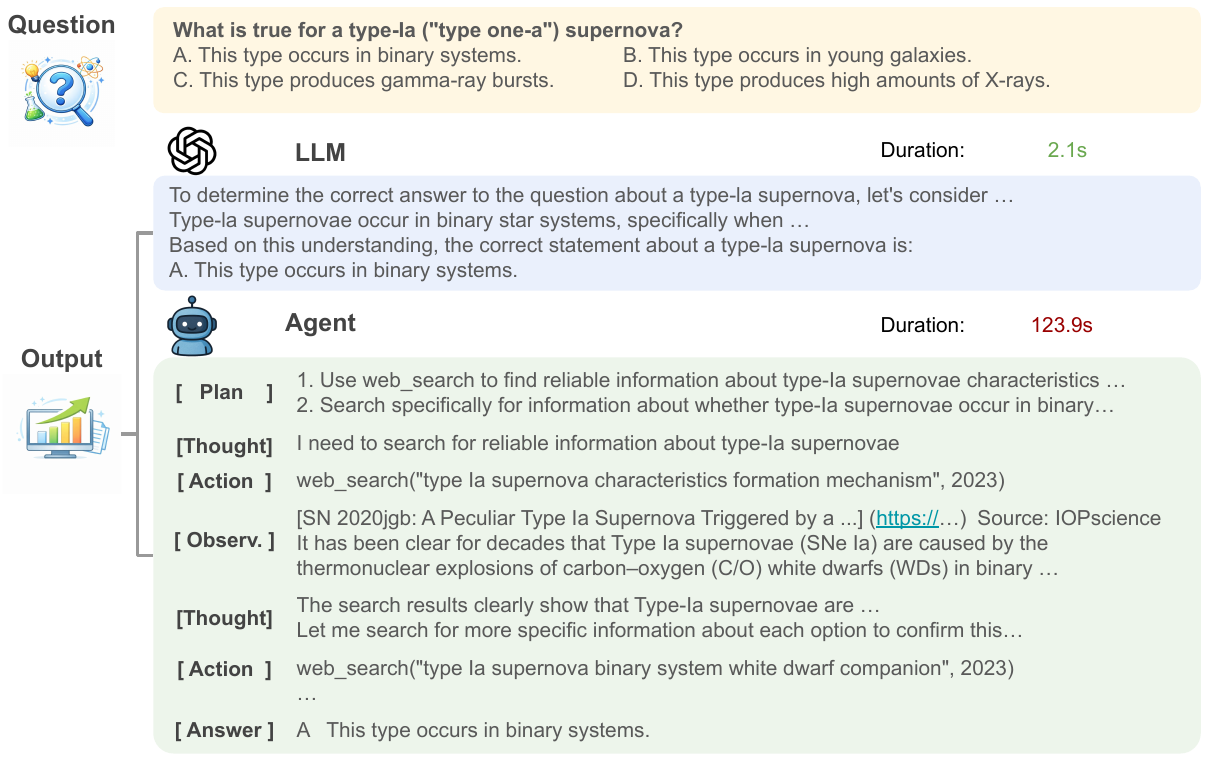}
    \caption{ Example illustrating routing between direct LLM inference and agent execution.
For this factual multiple-choice question, the LLM produces a correct answer quickly (2.1s), while the agent follows a multi-step search-and-reasoning process that is substantially slower (123.9s) but arrives at the same conclusion. This example highlights the accuracy–latency trade-off that motivates routing, where many queries fall within the capability boundary of direct LLM inference and do not require full agent execution.
    }
    \label{example}
\end{figure}
\newpage
\subsection{Method Details}
\begin{figure}[htbp]
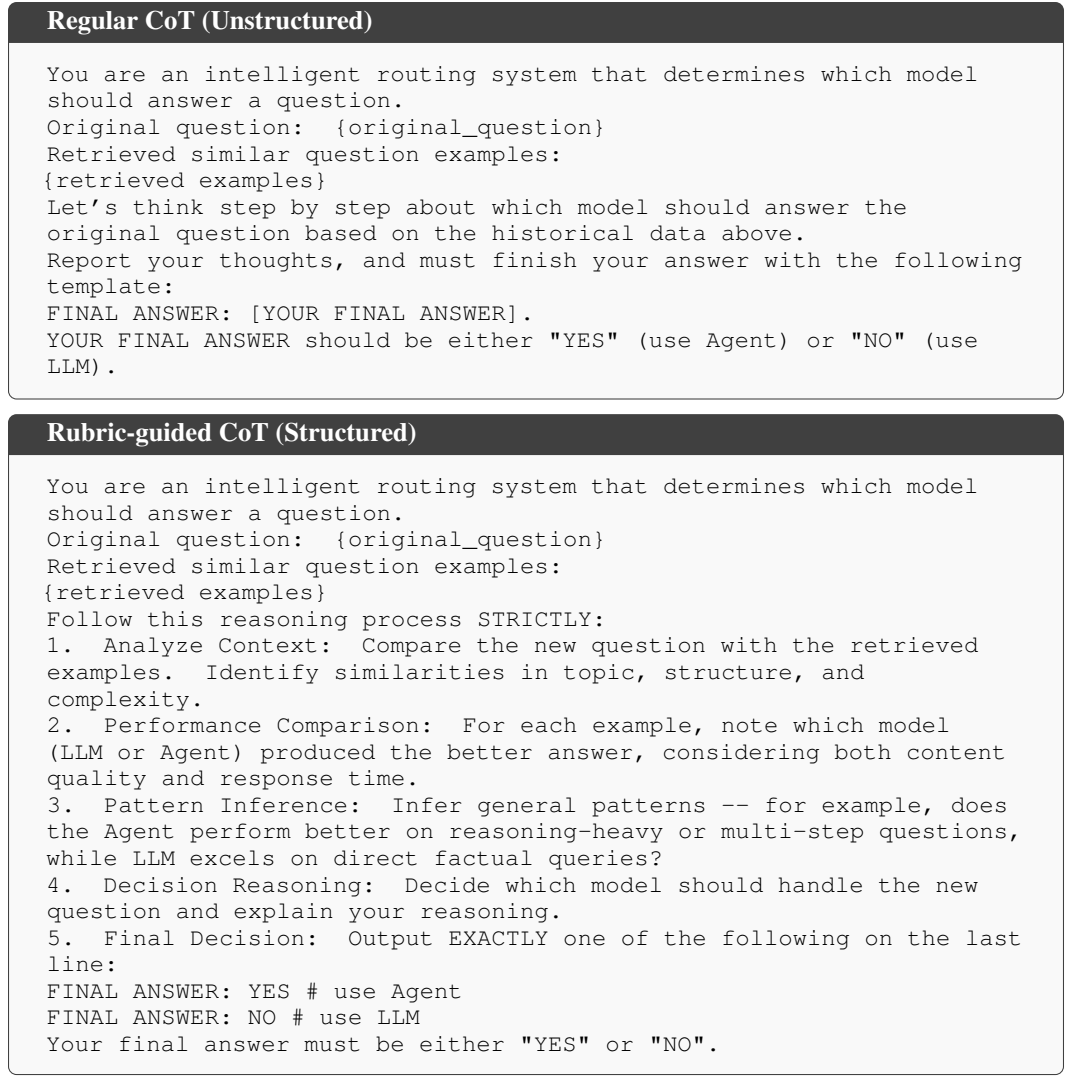

\centering


\begin{tcolorbox}[
  title=Regular CoT (Unstructured),
  colback=gray!5!white,
  colframe=gray!50!black,
  fonttitle=\bfseries,
  breakable,
  enhanced,
]
\small
\texttt{You are an intelligent routing system that determines which model should answer a question.}

\texttt{Original question: \{original\_question\}}

\texttt{Retrieved similar question examples:}

\texttt{\{retrieved examples\}}

\texttt{Let's think step by step about which model should answer the original question based on the historical data above.}

\texttt{Report your thoughts, and must finish your answer with the following template:}

\texttt{FINAL ANSWER: [YOUR FINAL ANSWER].}

\texttt{YOUR FINAL ANSWER should be either "YES" (use Agent) or "NO" (use LLM).}
\end{tcolorbox}
\begin{tcolorbox}[
  title=Rubric-guided CoT (Structured),
  colback=gray!5!white,
  colframe=gray!50!black,
  fonttitle=\bfseries,
  breakable,
  enhanced,
]
\small
\texttt{You are an intelligent routing system that determines which model should answer a question.}

\texttt{Original question: \{original\_question\}}

\texttt{Retrieved similar question examples:}

\texttt{\{retrieved examples\}}

\texttt{Follow this reasoning process STRICTLY:}

\texttt{1. Analyze Context: Compare the new question with the retrieved examples. Identify similarities in topic, structure, and complexity.}

\texttt{2. Performance Comparison: For each example, note which model (LLM or Agent) produced the better answer, considering both content quality and response time.}

\texttt{3. Pattern Inference: Infer general patterns --- for example, does the Agent perform better on reasoning-heavy or multi-step questions, while LLM excels on direct factual queries?}

\texttt{4. Decision Reasoning: Decide which model should handle the new question and explain your reasoning.}

\texttt{5. Final Decision: Output EXACTLY one of the following on the last line:}

\texttt{FINAL ANSWER: YES \# use Agent}

\texttt{FINAL ANSWER: NO \# use LLM}

\texttt{Your final answer must be either "YES" or "NO".}
\end{tcolorbox}
\vspace{0.3em}
\caption{Comparison between regular CoT and rubric-guided CoT prompts.}
\label{fig:Prompt Comparison}
\end{figure}
\subsection{Prompts for Ablation Study}
\label{app: ablation_prompt}
\begin{tcolorbox}[
  colback=gray!5!white,
  colframe=gray!50!black,
  title=Prompt Routing,
  fonttitle=\bfseries,
  breakable,
  enhanced,
  sharp corners=south,
  colbacktitle=gray!10!black
]
\label{Prompt Routing}
You are an intelligent routing system that determines which model should answer a question.

\textbf{Model A:}
\begin{itemize}
  \item Strengths: A fast, stable, general-purpose QA model that excels at natural language understanding, straightforward reasoning, and well-formatted outputs; ideal for simple to medium tasks.
\end{itemize}

\textbf{Model B:}
\begin{itemize}
  \item Strengths: A self-evolving reasoning agent capable of complex multi-step planning, self-consistency checking, and structured problem solving; slower but stronger in deep reasoning tasks.
\end{itemize}

Choose the most suitable model based only on these capability profiles and the question below.

\textbf{Question:} \{original\_question\}

Please answer only: \textbf{YES} (use Model B) or \textbf{NO} (use Model A).
\end{tcolorbox}

\begin{tcolorbox}[
  colback=gray!5!white,
  colframe=gray!50!black,
  title=Routing Prompt,
  fonttitle=\bfseries,
  breakable,
  enhanced,
  sharp corners=south,
  colbacktitle=gray!10!black
]
You are an intelligent routing system that determines which model should answer a question.

\textbf{Original question:} \{original\_question\}

\textbf{Retrieved similar question examples:}

\{chr(10).join(reference\_examples) if reference\_examples else ``No similar questions found''\}

Based on the similar questions and their historical performance, decide which model should answer the original question.

Please answer only: \textbf{YES} (use Model B) or \textbf{NO} (use Model A).
\end{tcolorbox}


\end{document}